\documentclass{article}
\usepackage{arxiv}  

\usepackage[utf8]{inputenc} 
\usepackage[T1]{fontenc}    
\usepackage{hyperref}       
\usepackage{url}            
\usepackage{booktabs}       
\usepackage{amsfonts}       
\usepackage{nicefrac}       
\usepackage{microtype}      
\usepackage{lipsum}         
\usepackage{graphicx}
\usepackage{natbib}
\usepackage{doi}

\usepackage{lineno,subcaption}
\usepackage{float}
\usepackage{multirow}
\usepackage{amssymb}
\usepackage{amsmath}
\usepackage{bbm}
\usepackage{amsthm}
\usepackage{cleveref}       
\usepackage{makecell}

\title{Bridging the Gap of AutoGraph between Academia and Industry:
    Analysing AutoGraph Challenge at KDD Cup 2020}

\date{}

\author{ Zhen Xu  \\
	4Paradigm, China\\
	\texttt{xuzhen@4paradigm.com} \\
	\And
	Lanning Wei \\
	4Paradigm, China\\
	Institute of Computing Technology,\\
	Chinese Academy of Sciences, China \\
	\texttt{weilanning@4paradigm.com} \\
	\And
	Huan Zhao \\
	4Paradigm, China\\
	\texttt{zhaohuan@4paradigm.com} \\
	\AND
    Rex Ying \\
	Stanford University, USA\\
	\texttt{rexying@stanford.edu} \\
	\And
	Quanming Yao \\
	Tsinghua University, China\\
	\texttt{qyaoaa@tsinghua.edu.cn} \\
	\And
	Wei-Wei Tu \\
	4Paradigm, China\\
	\texttt{tuweiwei@4paradigm.com} \\
	\And
	Isabelle Guyon \\
	ChaLearn, USA \\
	LISN/INRIA/CNRS, University Paris-Saclay, France\\
	\texttt{guyon@chalearn.org} \\
}

\renewcommand{\shorttitle}{AutoGraph Analyses}

\begin{document}
\maketitle

\begin{abstract}
	Graph structured data is ubiquitous in daily life and scientific areas and has attracted increasing attention. Graph Neural Networks (GNNs) have been proved to be effective in modeling graph structured data and many variants of GNN architectures have been proposed. However, much human effort is often needed to tune the architecture depending on different datasets. Researchers naturally adopt Automated Machine Learning on Graph Learning, aiming to reduce the human effort and achieve generally top-performing GNNs, but their methods focus more on the architecture search. To understand GNN practitioners' automated solutions, we organized AutoGraph Challenge at KDD Cup 2020, emphasizing on automated graph neural networks for node classification. We received top solutions especially from industrial tech companies like Meituan, Alibaba and Twitter, which are already open sourced on Github. After detailed comparisons with solutions from academia, we quantify the gaps between academia and industry on modeling scope, effectiveness and efficiency,  and show that (1) academia AutoML for Graph solutions focus on GNN architecture search while industrial solutions, especially the winning ones in the KDD Cup, tend to obtain an overall solution (2) by neural architecture search only, academia solutions achieve on average 97.3\% accuracy of industrial solutions (3) academia solutions are cheap to obtain with several GPU hours while industrial solutions take a few months' labors. Academic solutions also contain much fewer parameters.
\end{abstract}


\section{Introduction}

Graph structured data has been prominent in our life and various tasks are studied based upon, including recommendation on Social Networks \citep{19_fan_graph_nerual}, traffic forecasting on road networks \citep{18_li_diffusion_comvolutional}, drug discovery on molecule graph \citep{19_torng_graph_convolutional}, link prediction on knowledge graph \citep{20_zhang_relational_graph}, etc. Graph Neural Networks (GNN) \citep{17_kipf_semi_supervised} have been proved to be effective in modeling graph data and tremendous GNN architectures are proposed every year \citep{hamilton2017inductive,xu2018how,wu2019simplifying,velickovic2018graph}.

When applying GNN on graph structured data, expertise and domain knowledge is often required and numerous human effort is required to adapt to new datasets. Automated Machine Learning (AutoML) \citep{hutter2019automated,yao2018taking} aims to reduce human efforts in deploying on various applications. AutoML, especially Neural Architecture Search (NAS), has been successfully explored on tremendous applications, including Image Classification \citep{tan2019efficientnet}, Object Detection \citep{20tanefficientdet}, Semantic Segmentation \citep{Nekrasov2019fast}, Language Modeling \citep{jiang2019improved}, Time Series Forecasting \citep{chen2021scale}, etc. As a result, researchers start to explore Automated Graph Neural Networks (AutoGraph). AutoGraph researchers focus mainly on automatically designing GNN architectures by NAS. The majority of these methods focus on designing the aggregation functions/layers in GNNs with different search algorithms \citep{gao2019graphnas,zhou2019auto,yoon2020autonomous,li2021one,peng2019learning}. Other works, SANE~\citep{zhao2021search} and 
AutoGraph~\citep{li2020autograph}, provide the extra layer-wise skip connections design dimension; GNAS~\citep{cai2021rethinking}, 
DeepGNAS~\citep{feng2021search} and Policy-GNN~\citep{lai2020policy} learn to design the depth of GNNs. DiffMG \citep{ding2021diffmg} proposed to use NAS to search data-specific meta-graphs in heterogeneous graph, and PAS \citep{wei2021pooling} is proposed to search data-specific pooling architectures for graph classification. The recently proposed F$^2$GNN~\citep{wei2021designing} method decouples the design of aggregation operations with architecture topology, which is not considered before.

Despite the rich literature from academia, we ask the question of how AutoGraph is used in industrial practitioners. Towards this end, we organize the first AutoGraph challenge at KDD Cup 2020, collaborated with 4Paradigm, ChaLearn and Stanford University. This challenge asks participants to provide AutoGraph solutions for node classification task. The code is executed by the platform on various graph datasets without any human intervention. Through the AutoGraph challenge, we wish to push forward the limit of AutoGraph as well as  to understand the gap between industrial solutions and academia ones. In this paper, we first introduce the AutoGraph challenge setting. Then, we present the winning solutions which are open sourced for everyone to use. At last, we experiment further and compare with NAS methods for GNN methods and quantify empirically the gap with respect to top solutions. \textbf{We conclude three gaps of AutoGraph between academia and industry: Modeling scope, Effectiveness and Efficiency.}

\section{Challenge background}

\subsection{General statistics}

The AutoGraph challenge lasted for two months. We received over 2200 submissions and more than 140 teams from both high-tech companies (Ant Financial, Bytedance, Criteo, Meituan Dianping, Twitter, NTT DOCOMO, etc.) and universities (MIT, UCLA, Tsinghua University, Peking University, Nanyang Technological University, National University of Singapore, IIT Kanpur, George Washington University, etc.), coming from various countries. The top three teams are: \texttt{aister, PASA\_NJU, qqerret}. 
Top 10 winners' information are shown in Table~\ref{tab:winners}. 
The 1st winner \texttt{aister} comes from Meituan Dianping, a company on location-based shopping and retailing service. This makes the challenge particularly valuable since we can compare academic solutions with industrial best AutoGraph practices.

\begin{table}[ht]
    \centering
    \caption{General information about winning teams. Two teams tie on the 6th and 10th place. We list them both.}
    \label{tab:winners}
    \begin{tabular}{lllclll}
    \toprule
        Place & Team Name & Institute & & Place & Team Name & Institute \\
        \midrule
        1st & \texttt{aister} & Meituan Dianping & & &  \\
        2nd  & \texttt{PASA\_NJU} & Nanjing University &  & 6th & \makecell[l]{\texttt{SmartMN-THU}\\\texttt{JunweiSun}} & \makecell[l]{Tsinghua University\\Beijing University of Posts\\and Telecommunications} \\
        3rd  & \texttt{qqerret} & Ant Financial & & 8th & \texttt{u1234x1234} & self-employed \\
        4th  & \texttt{common} & Alibaba Inc. & & 9th & \texttt{AML} & Ant Financial \\
        5th  & \texttt{PostDawn} & Zhejiang University & & 10th & \makecell[l]{\texttt{supergx}\\\texttt{daydayup}} & \makecell[l]{Nanyang Tech. University\\Hikvision Inc.} \\
    \bottomrule
    \end{tabular}
\end{table}

\subsection{Problem formulation} The task of AutoGraph challenge is node classification under the transductive setting. Formally speaking, consider a graph $\mathcal G = (\mathcal{V}, \mathcal{E})$, where $\mathcal{V} = \{v_1,\cdots, v_N\}$ is the set of nodes, i.e. $|\mathcal{V}|=N$ and $\mathcal{E}$ is the set of edges, which is usually encoded by an adjacency matrix $A \in [0,1]^{N \times N}$. $A_{ij}$ is positive if there is an edge connecting from node $v_i$ to node $v_j$. Additionally, a feature matrix $X \in \mathbb{R}^{N \times D}$ gives features of each node. Each node $v_i$ has a class label $y_i \in \mathcal{L}=\{1, \cdots, c\}$, resulting in the label vector $Y \in \mathcal{L}^N$. In the transductive semi-supervised node classification task, part of labels are available during training and the goal is to learn a mapping $\mathcal F: \mathcal V \rightarrow \mathcal L$ and predict classes of unlabeled nodes.

\subsection{Protocol} 
The protocol of AutoGraph challenge is straightforward. Participants should submit a python file containing a \textsf{Model} class with required \textsf{fit} and \textsf{predict} method. 
We prepare an ingestion program reading dataset and instantiate the class and call \textsf{fit} and \textsf{predict} method until prediction finishes or the running time has reached the budget limit.  Ingestion program outputs model's prediction on test data and save to a shared space. 
Then, 
another scoring program reads the prediction and ground truth and outputs evaluation scores. 
The execution of the program is totally on the challenge platform. When developing locally, 
we provide script to call \textsf{model.py} file methods directly.

\subsection{Metric} We use \textsf{Accuracy (Acc)} and \textsf{Balanced Accuracy (BalAcc)} as evaluation metrics, defined as 
\begin{equation*}
\textsf{Acc} = \frac{1}{|\Omega|} 
\sum\nolimits_{i \in \Omega} \mathbbm{1}_{\hat y_i = y_i},
\qquad
\textsf{BalAcc} = \frac{1}{|C|} 
\sum\nolimits_{i \in C} \textsf{Recall}_i,
\end{equation*}
where $\Omega$ is the set of test nodes indexes, $y_i$ is the ground truth label for node $v_i$ and $\hat y_i$ is the predicted label, 
$C$ is the set of classes and $\textsf{Recall}_i$ is the recall score for class $i$. \textsf{Accuracy (Acc)} is used in the challenge to rank participants and \textsf{Balanced Accuracy (BalAcc)} is applied for additional analyses since it takes into account the imbalanced label distribution of datasets.

\subsection{Datasets} Fifteen graph datasets were used during the competition: 5 public datasets were directly downloadable by the participants so they could develop their solutions offline. 
Five feedback datasets were made available on the platform during the feedback phase to evaluate AutoGraph algorithms on the public leaderboard. Finally, the AutoGraph algorithms were evaluated with 5 private datasets, without human intervention. 
These dataset are quite diverse in domains, shapes, density and other graph properties because we expect AutoGraph solutions to have good generalization ability.
On the other hands, we intentionally keep the characteristics of 5 feedback datasets and 5 private datasets similar to enable transferability. 
We summarize dataset statistics in Table \ref{tab:stats}. The licenses and original sources of these datasets are also provided\footnote{\url{https://github.com/AutoML-Research/AutoGraph-KDDCup2020}}.

\begin{table}[ht!]
\caption{{\bf Statistics of all datasets}. ``Avg Deg'' is the average number of edges per node. ``Directed'' and ``Weighted'' indicate the two properties of a graph. ``Skewness'' here is calculated by number of nodes in the largest class divided by number of nodes in the smallest class.}
\label{tab:stats}
\centering
\resizebox{\textwidth}{!}{
\begin{tabular}{cllrrrrrccr} 
\toprule
Dataset & Phase & Domain & \#Node & \#Edge & \#Feature & \#Class & Avg Deg & Directed? & Weighted? & Skewness \\
\midrule
 a & Public  & Citation  & 2.7K & 5.3K  & 1.4K  & 7 & 1.9 & F & F & 5 \\
 b & Public  & Citation  & 3.3K & 4.6K  & 3.7K  & 6 & 1.4 & F & F & 3 \\
 c & Public  & Social  & 10K & 733K  & 0.6K  & 41 & 73.3 & F & F & 81 \\
 d & Public  & News  & 10K & 2,917K & 0.3K  & 20 & 291.7 & T & T & 467 \\
 e & Public  & Finance  & 7.5K & 7.8K & 0  & 3 & 1.0 & F & F & 111  \\
\midrule
 f & Feedback & Sales & 10K & 194K & 0.7K  & 10  & 19.4 & F & F & 18  \\
 g & Feedback & Citation & 10K & 41K  & 8K  & 5  & 4.1 & F & F & 6   \\
 h & Feedback & Medicine & 10K & 2,461K & 0.3K  & 23  & 246.1 & T & T & 1,773 \\
 i & Feedback & Finance & 15K & 16K & 0  & 3   & 1.1 & F & F & 213  \\
 j & Feedback & Medicine & 11K & 22K & 0  & 9  & 2.0 & F & F & 227  \\
\midrule
 k  & Private  & Sales & 8K  & 119K   & 0.7K   & 8 & 14.9 & F & F & 6  \\
 l  & Private  & Citation & 10K  & 40K  & 7K  & 15  & 4 & F & F & 34  \\
 m  & Private  & News  & 10K  & 1,425K & 0.3K  & 8  & 142.5 & T & T & 360  \\
 n  & Private  & Finance & 14K   & 22K & 0  & 10   & 1.6 & F & F & 61  \\
 o  & Private  & Social & 12K     & 19K   & 0  & 19 & 1.6 & F & F & 62   \\
\bottomrule
\end{tabular}
}
\vspace{0.5cm}
\end{table}

\section{Solutions}

In this part, we introduce various methods suitable for the AutoGraph challenge, including the provided challenge baseline and solutions from top-3 winners. We conclude the first gap at the end.

\textbf{Baselines (GCN(L2)).} In the provided baselines, there is no feature engineering except for using the raw node features. For graph without node features, e.g. dataset i,j, one hot encoding is used to unroll the node lists to a dummy feature table. During model training, a MLP is first used to map node features to the same embedding dimension. Then a two layer vanilla GCN is applied for learning node embeddings. Another MLP with softmax outputs the final classification. Dropout is used. All the hyperparameters are fixed by experience. No time management since the model is simple and one full training will not cost more than the allowed time budget.

\textbf{1st placed winner.} The 1st winner is from team \texttt{aister}. Their code is open source here\footnote{\url{https://github.com/aister2020/KDDCUP_2020_AutoGraph_1st_Place}}. The authors use four GNN models, two spatial ones: GraphSage \citep{hamilton2017inductive} and GAT \citep{velickovic2018graph}, two spectral ones: GCN \citep{17_kipf_semi_supervised} and TAGConv \citep{du2017topology} to process node features collectively. For each GNN model, a heavy search is applied offline to determining the important hyperparameters as well as the boundaries. In the online stage, they use a smaller search space to determine the hyperparameters. In order to accelerate the search, they do not fully train each configuration but instead early stop in 16 epochs if the validation loss is not satisfactory. Additional features are used: node degrees, distribution of 1-hop and 2-hop neighbor nodes' features, etc.

\textbf{2nd place winner.} The 2nd winner is from team \texttt{PASA\_NJU}. Their code is open source here\footnote{\url{https://github.com/Unkrible/AutoGraph2020}}. They also split the solution in two stages: offline stage and online stage. In the offline stage, the authors train a decision tree based on public data and other self collected datasets to classify graph type into one of three classes. Then they use GraphNAS \citep{gao2019graphnas} to search massively optimal GNN architectures including aggregation function, activation, number of heads in attention, hidden units, etc. In the online stage, the authors rapidly classify the dataset and fine tune the offline searched model.

\textbf{3rd place winner.} The 3rd winner is from team 
\texttt{qqerret}. Their code is open source here\footnote{\url{https://github.com/white-bird/kdd2020_GCN}}. The core model is a variant of spatial based GNN, which aggregates 2-hop neighbors of a node with additional linear parts for the node itself. Basically, the new embedding of node $i$ is $\hat h(i) = \sum_{j \in N_2(i)} a_j h(j) + \alpha (wh(i) + b)$. Additionally, in the GNN output layer, a few features per node are concatenated for final fully connected layer, including number of edges, whether this node connects to a central node who has a lot of edges, label distribution of 1-hop neighbor nodes, and label distribution of 2-hop neighboring nodes.

\begin{figure}
    \vspace{0.5cm}
    \centering
    \includegraphics[width=1\textwidth]{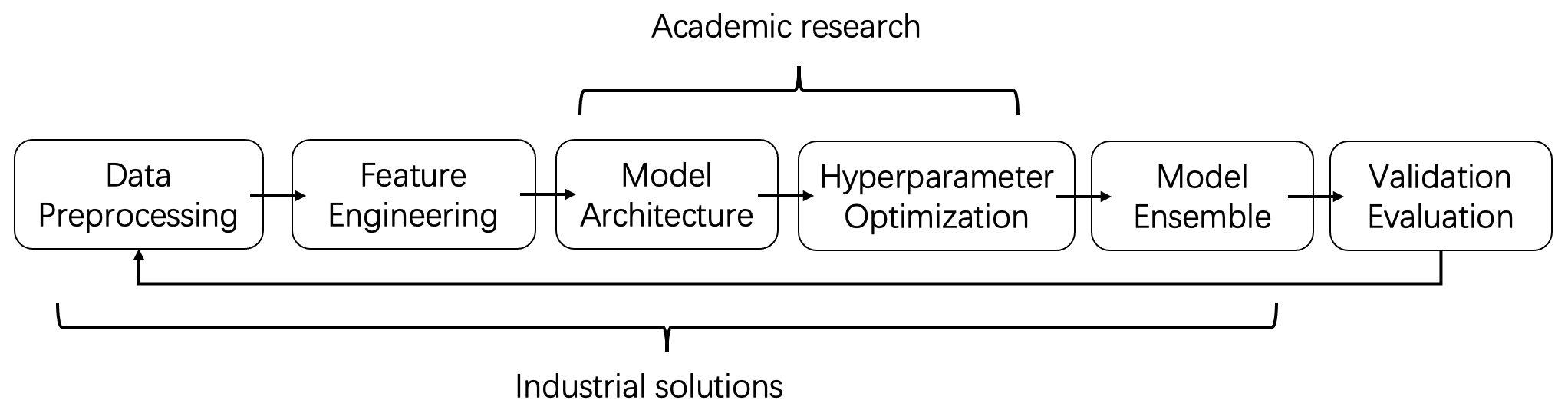}
    \caption{Illustraion of AutoGraph scope. Industrial people provide a full pipeline solution that covers from data preprocessing to evaluation. Academic researchers focus mainly on model architecture and hyperparameter optimization.}
    \label{fig:scope}
    \vspace{0.5cm}
\end{figure}

\section{Results}

We conduct additional experiments after the AutoGraph challenge to 
further analyze the results. 
We first reproduce winning solutions and then we compare with academia solutions. 
Three gaps are concluded. The first gap is presented as follows and two other gaps are concluded in Sec 4.2.

\textbf{Gap \#1: Modeling scope is the first gap of AutoGraph between academia and industry.} In academia, researchers focus mainly on Neural Architecture Search methods to find better GNN architectures. Their contributions differ in their search space, search strategy and evaluation methods. However,  industrial solutions, e.g. 1st solution, focus more on the feature engineering and model ensemble. For GNN architectures, they prefer existing found ones with little modification. In other words, industrial people provide a full pipeline solution including data preprocessing, feature engineering, model architecture, hyperparameter optimization, model ensemble, while academia researchers focus on model architecture part only. The gap is also illustrated in Figure~\ref{fig:scope}. It might be an interesting direction for both groups to merge, i.e. AutoGraph researchers could explore the automated feature engineering, automated ensemble and AutoGraph practitioners could adopt NAS methods for GNN.

\subsection{Reproducing winning solutions}

We reproduce all winning methods on all the datasets and include their results in Table~\ref{tab:reprod}. We observe that all three winning solutions are close in performance and all significantly beat the GCN baselines. 
On the other hand, in the AutoGraph challenge, 
due to the nature of the competition, 
we rank methods based on accuracy, we state that this is not sufficient to evaluate comprehensively solutions from the scientific perspective. We add balanced accuracy here just to show that for some methods that show close performance in accuracy, they could diverge a lot in balanced accuracy. Regarding both accuracy and balanced accuracy, we conclude that 1st solution which comes from Meituan Dianping Company, is indeed best among top winners. Thus, we will later use their solutions for comparing with academia solutions. These winning solutions are already open sourced, which are reproducible and lower the barriers of using AutoGraph.

\begin{table}[H]
\caption{  Accuracy and Balanced accuracy of top methods on all datasets (\%). Baseline is a two layer GCN.}
\label{tab:reprod}
\centering
\begin{tabular}{clcccccccc}
\toprule
Dataset & Phase & \multicolumn{2}{c}{Baseline (GCN(L2))} & \multicolumn{2}{c}{1st solution} & \multicolumn{2}{c}{2nd solution} & \multicolumn{2}{c}{3rd solution} \\
\midrule
& & \textsf{Acc} & \textsf{BalAcc} & \textsf{Acc} & \textsf{BalAcc} & \textsf{Acc} & \textsf{BalAcc} & \textsf{Acc} & \textsf{BalAcc} \\

\midrule
 a & Public & 85.7 & 84.9 & {\bf 88.5} & {\bf 87.8} & 88.2 & 87.2 & 87.2 & 85.5 \\
 b & Public & 71.4 & 67.8 & 75.2 & {\bf 71.2} & {\bf 75.8} & {\bf 71.2} & 75.6 & 69.0 \\
 c & Public & 86.5 & 72.0 & 94.3 & 87.5 & 94.2 & 90.9 & {\bf 95.4} & {\bf 91.3} \\
 d & Public & 93.7 & 6.1  & {\bf 96.5} & {\bf 48.7} & 95.1 & 28.8 & 94.6 & 21.0 \\
 e & Public & 59.6 & 38.8 & 88.7 & {\bf 92.8} & 88.5 & 90.7 & {\bf 88.8} & {\bf 92.8} \\
\midrule
 f & Feedback & 86.6 & 78.2 & {\bf 92.8} & {\bf 92.1} & 92.3 & {\bf 92.1} & 92.4 & 91.4 \\
 g & Feedback & 94.7 & 92.8 & 95.3 & 93.5 & 95.6 & 93.8 & {\bf 95.8} & {\bf 94.2} \\
 h & Feedback & 90.4 & 8.8  & {\bf 93.5} & {\bf 26.3} & 92.2 & 17.6 & 92.1 & 16.6 \\
 i & Feedback & 88.2 & 59.2 & 88.4 & 87.5 & 88.4 & {\bf 92.6} & {\bf 88.5} & 91.1 \\
 j & Feedback & 90.7 & 68.1 & 95.9 & 89.0 & 96.1 & {\bf 93.7} & {\bf 96.6} & 93.3 \\
\midrule
 k & Private & 93.5 & 92.2 & {\bf 95.5} & {\bf 94.4} & {\bf 95.5} & {\bf 94.4} & 94.8 & 93.1 \\
 l & Private & 90.9 & 84.5 & {\bf 94.9} & {\bf 92.6} & 94.7 & 91.8 & 94.5 & {\bf 92.6} \\
 m & Private & 85.5 & 24.5 & {\bf 98.1} & {\bf 79.7} & 95.7 & 69.0 &  98.0 & 79.4 \\
 n & Private & 85.6 & 47.3 & {\bf 99.0} & 97.3 & {\bf 99.0} & {\bf 98.4} & 98.9 & 97.0 \\
 o & Private & 49.6 & 15.6 & 91.0 & 84.6 & 91.3 & {\bf 90.6} & {\bf 91.4} & 88.5 \\
\bottomrule
\end{tabular}
\end{table}

\subsection{Neural Architecture Search for GNN}

We further adopt NAS methods for GNN and compare with the baseline and 1st solution coming from industry. We choose the recent F$^2$GNN \citep{wei2021designing}, 
which searches for data-specific GNN topology, in our experiment. 
To compare fairly with GCN baselines, we fix the aggregation to GCN and search only the GNN topology, 
which we call F$^2$GCN. Since F$^2$GCN requires at least 4 layers, we also run a 4 layer GCN baseline for better comparison. The results are given in Table \ref{tab:nas}.

\begin{minipage}{\textwidth}
    \vspace{0.5cm}
  \begin{minipage}[b]{0.55\textwidth}
    \centering
    \scalebox{0.9}{
    \begin{tabular}{cllrr}
    \toprule
         Dataset & GCN(L2) & GCN(L4) & F$^2$GCN(L4) & 1st solution \\
         \midrule
         a & 85.7 & 84.4 & 84.4 (95.4) & 88.5 (100)  \\
         b & 71.4 & 70.5 & 71.3 (94.8) & 75.2 (100) \\
         c & 86.5 & 82.3 & 92.8 (98.4) & 94.3 (100)  \\
         d & 93.7 & 93.6 & 93.9 (97.3) & 96.5 (100)  \\
         e & 59.6 & 87.5 & 88.4 (99.7) & 88.7 (100)  \\
         \midrule
         f & 86.6 & 87.6 & 92.1 (99.2) & 92.8 (100)  \\
         g & 94.7 & 93.4 & 95.3 (100)  & 95.3 (100)  \\
         h & 90.4 & 90.3 & 90.1 (96.4) & 93.5 (100)  \\
         i & 88.2 & 87.6 & 88.3 (99.9) & 88.4 (100)  \\
         j & 90.7 & 83.6 & 95.3 (99.4) & 95.9 (100)  \\
         \midrule
         k & 93.5 & 93.2 & 93.4 (97.9) & 95.5 (100)  \\
         l & 90.9 & 89.1 & 92.9 (97.9) & 94.9 (100)  \\
         m & 85.5 & 86.1 & 86.1 (87.8) & 98.1 (100)  \\
         n & 85.6 & 95.2 & 96.7 (97.7) & 99.0 (100)  \\
         o & 49.6 & 71.8 & 88.8 (97.6) & 91.0 (100)  \\
         \midrule
         Avg &  &  &   - (97.3) &    - (100) \\
    \bottomrule
    \end{tabular}
      }
      \captionof{table}{Accuracy comparison of GCN baselines, F$^2$GCN and industrial best solution (\%). L2, L4 means 2 and 4 layers for the GNN architecture. Numbers in parentheses are relative accuracy w.r.t 1st solution. We regard 1st solution as 100\%. Last line is the average percentage.}
      \label{tab:nas}
    \end{minipage}
    \hspace{0.2cm}
    \begin{minipage}[b]{0.4\textwidth}
        \centering
        \includegraphics[width=0.8\textwidth]{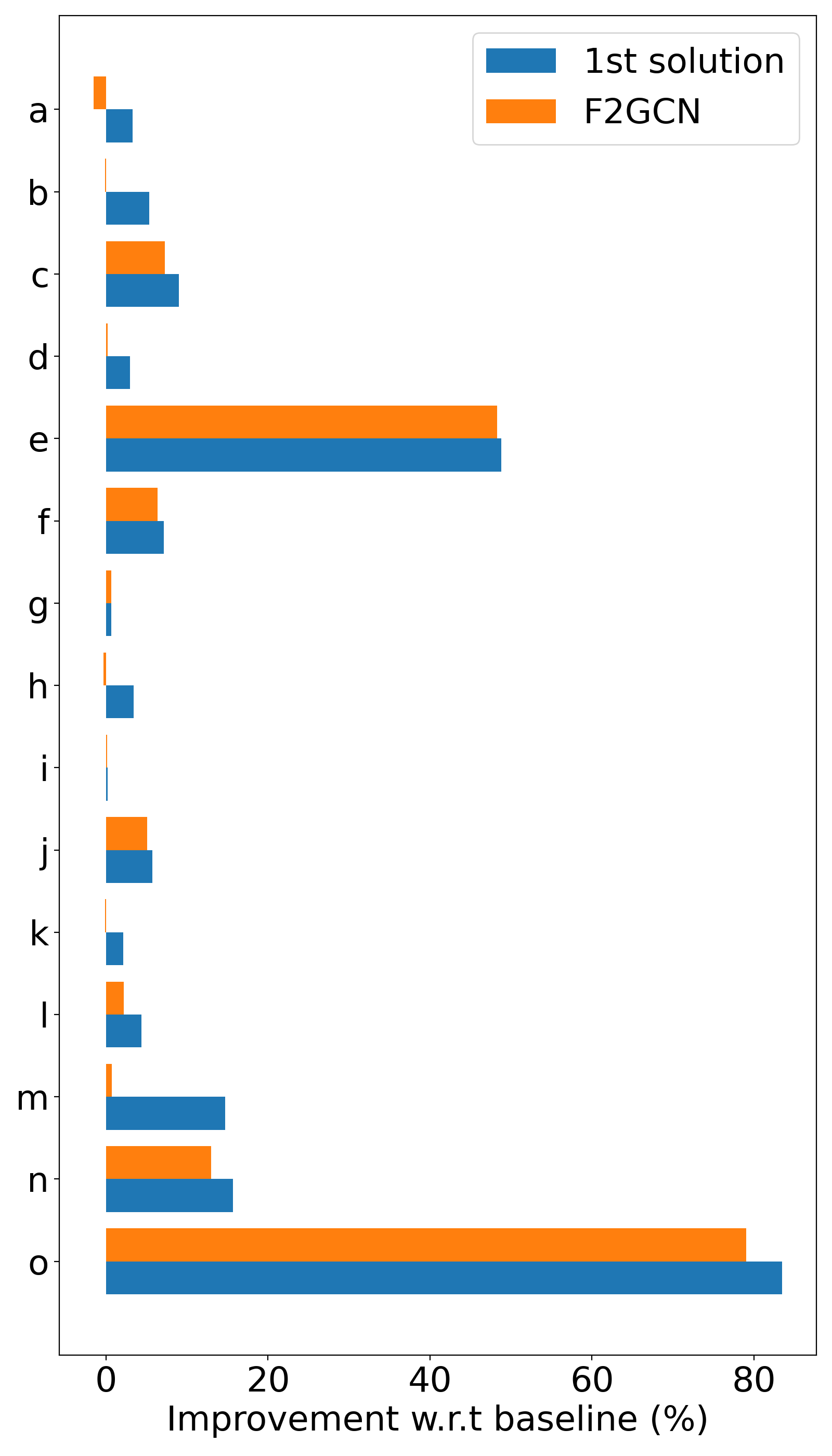}
        \captionof{figure}{Accuracy improvement with respect to baseline.}
        \label{fig:nas}
    \end{minipage}
    \vspace{1cm}
\end{minipage}

\textbf{Gap \#2: Effectiveness is the second gap of AutoGraph between academia and industry.}  We observe from  Table \ref{tab:nas} and Figure~\ref{fig:nas} that all baselines and F$^2$GCN methods are not as good as 1st winning solution. However, for many datasets, e.g. e, f, g, i, j, F$^2$GCN is very close to the best industrial solution. On average, F$^2$GCN which focuses only on architecture search, reaches 97.3\% of best solution. Note that the 1st solution constructs additional node features and uses multiple GNN architectures for ensemble while F$^2$GCN does not use any feature engineering or model ensemble. This shows the effectiveness of winner's engineering practices as well as F$^2$GCN's adaptive NAS search. Winning teams also have access to public datasets and public leaderboard to iteratively fine tune their methods. F$^2$GCN does not assume any prior knowledge of the datasets, which shows further its effectiveness.

To better understand the solutions, 
we calculate the number of parameters of baseline, F$^2$GCN, and 1st solution, 
as shown in Table \ref{tab:param}.

\textbf{Gap \#3: Efficiency is the third gap of AutoGraph between academia and industry.} From Table \ref{tab:param} and Figure~\ref{fig:param},  F$^2$GCN uses significantly fewer parameters than the best industrial solution on most datasets (13 out of 15). 
On average,  F$^2$GCN consumes 45.1\% of the 1st solution in terms of parameter size, 
which is quite resource efficient. Note that feature engineering and ensemble do not contain additional parameters and basically, F$^2$GCN searches one GNN model to compete with ensemble of 4 types of GNN models in 1st solution. 
As for time devotion,  winning solutions come from a team's months of work, which consists of 5 or more members. F$^2$GCN only runs for a few GPU hours per dataset, demonstrating its time efficiency compared to industrial solutions.

\begin{minipage}{\textwidth}
    \hspace{-0.4cm}
  \begin{minipage}[b]{0.45\textwidth}
    \centering
    \scalebox{0.95}{
    \begin{tabular}{clrr}
    \toprule
         Dataset & GCN(L2) & F$^2$GCN(L4) & 1st solution \\
         \midrule
         a & 0.023  & 0.908 (75.7)& 1.199 (100) \\
         b & 0.059  & 0.700 (44.2)& 1.583 (100) \\
         c & 0.011  & 1.598 (98.0)& 1.631 (100) \\
         d & 0.006  & 0.042 (3.20)& 1.296 (100) \\
         e & 0.121  & 0.354 (31.8)& 1.114 (100) \\
         \midrule
         f & 0.013  & 0.039 (2.30)& 1.688 (100) \\
         g & 0.134  & 0.313 (13.1)& 2.389 (100) \\
         h & 0.006  & 0.271 (20.9)& 1.294 (100) \\
         i & 0.241  & 2.269 (113.0) & 2.013 (100) \\
         j & 0.171  & 0.834 (60.6)& 1.376 (100) \\
         \midrule
         k & 0.012 & 1.478 (108.0)  & 1.395 (100) \\
         l & 0.108  & 0.614 (25.6)& 2.395 (100) \\
         m & 0.005  & 0.010 (0.80)& 1.278 (100) \\
         n & 0.218  & 0.488 (27.8)& 1.756 (100) \\
         o & 0.192  & 0.822 (52.5)& 1.565 (100) \\
         \midrule
         Avg &   & - (45.1) & - (100)\\
    \bottomrule
    \end{tabular}
      }
      \captionof{table}{Number of parameters of baseline, 1st solution and F$^2$GCN (Unit: Millions). Numbers in parentheses are relative \# parameters w.r.t 1st solution. We regard 1st solution as 100\%. Last line is the average percentage. }
      \label{tab:param}
    \end{minipage}
    \hspace{0.3cm}
    \begin{minipage}[b]{0.5\textwidth}
        \centering
        \includegraphics[width=1.1\textwidth]{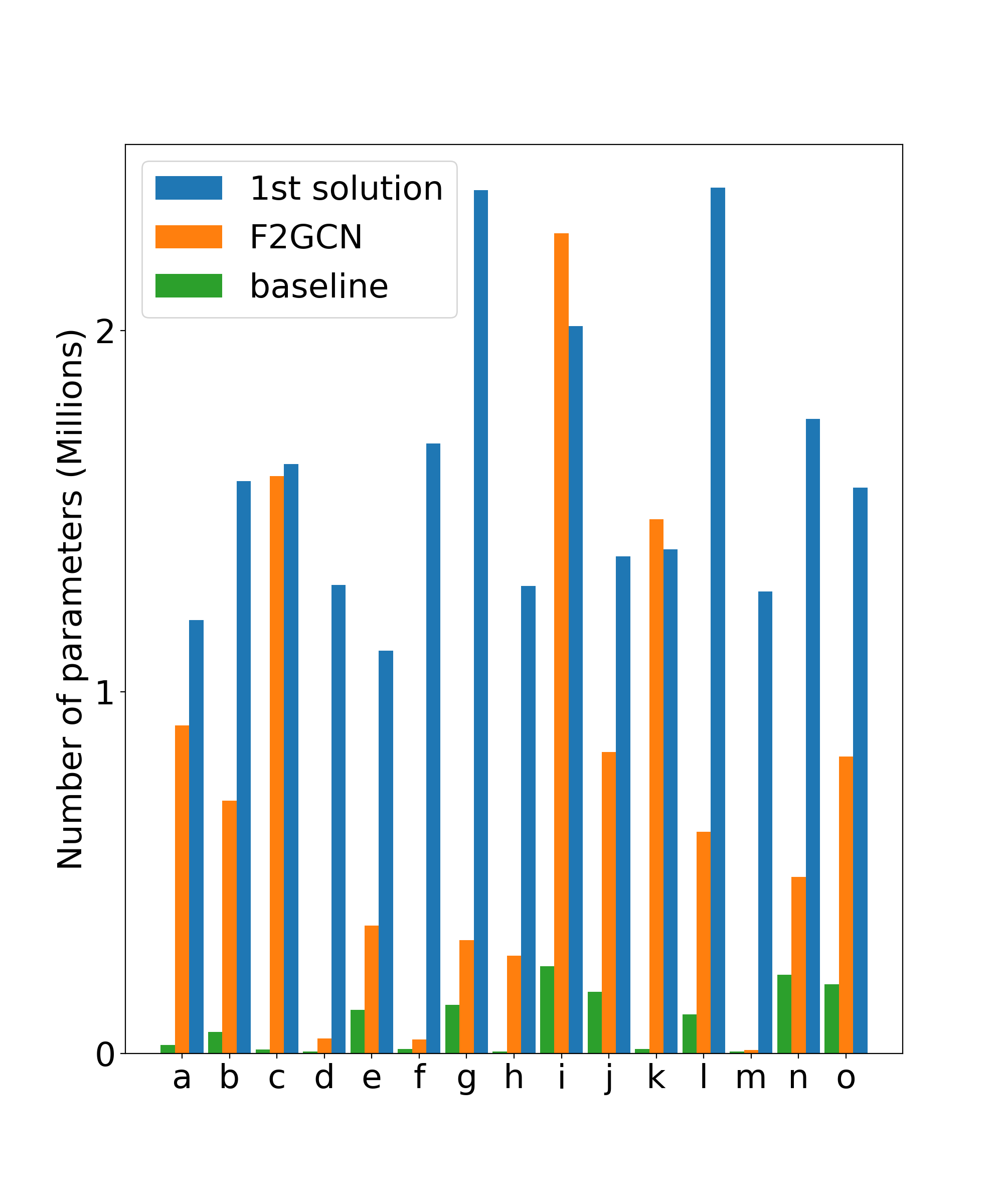}
        \captionof{figure}{Comparison on number of parameters of baseline, 1st solution and F$^2$GCN.}
        \label{fig:param}
    \end{minipage}
    \vspace{1cm}
\end{minipage}

\section{Conclusion}

We organized the first Automated Graph Learning (AutoGraph) Challenge at KDD Cup 2020. We presented in this paper its settings, dataset, and solutions, which are all open sourced. We ran additional post-challenge experiments to compare the baseline (Graph Convolution Network (GCN)), the winning solution (feature engineering-based ensemble of various Graph Neural Networks), and a recent and efficient Neural Architecture Search (NAS) for GNN method called F$^2$GCN. This paper provides results that could bridge the gap between academic research and industry practices, by correcting bias favoring certain approaches. This gap is currently at 3 aspects: {\bf Gap \#1 modeling scope.} (academia focuses more on model-centric approaches, emphasizing NAS; industry emphasizes data centric approaches and feature engineering); {\bf Gap \#2 effectiveness.} (academic solutions are perceived by industry to be less effective than their industry counterpart);  {\bf Gap \#3  efficiency.} (academic solutions are perceived to be parsimonious or slower than industry solutions).
Our results indicate that the ``academic'' NAS-based approach that we applied attains performances closely matching those of the winning industry solution, while being both faster and more parsimonious in number of parameters, therefore closing 
Gap~\#2 and \#3. Moreover,  we hope that these results will help reducing Gap~\#1, 
by encouraging industry practitioners to apply NAS methods (and particularly F$^2$GCN), eventually combining the best of both approaches. We believe the results we obtained are significant, since they involve a benchmark on 15 datasets.

\section*{Acknowledgments}

Funding and support have been received by several research grants, including 4Paradigm, Big Data Chair of Excellence FDS Paris-Saclay, Paris Région Ile-de-France, and ANR Chair of Artificial Intelligence HUMANIA ANR-19-CHIA-0022, ChaLearn, Microsoft, Google. We acknowledge the following people for helping organize AutoGraph challenge: Xiawei Guo, Shouxiang Liu. We also appreciate the following people and institutes for open sourcing datasets which are used in our use cases: Andrew McCallum, C. Lee Giles, Ken Lang, Tom Mitchell, William L. Hamilton, Maximilian Mumme, Oleksandr Shchur,  David D. Lewis, William Hersh, Just Research and Carnegie Mellon University, NEC Research Institute, Carnegie Mellon University, Stanford University, Technical University of Munich, AT\&T Labs, Oregon Health Sciences University.

\bibliographystyle{unsrtnat}
\bibliography{ref}  






\end{document}